\documentclass[runningheads]{llncs}

\usepackage[T1]{fontenc}

\usepackage{graphicx}
\usepackage{amssymb}
\usepackage{amsmath}
\usepackage[export]{adjustbox} 
\usepackage{booktabs}
\usepackage{xcolor}
\usepackage{hyperref}
\hypersetup{
    colorlinks=true,
    linkcolor=blue,      
    citecolor=red,       
    urlcolor=cyan,       
}

\begin{document}
\title{HULFSynth: An INR based Super-Resolution and Ultra Low-Field MRI Synthesis via Contrast factor estimation}
\titlerunning{HULFSynth}

\author{Pranav Indrakanti \orcidID{0009-0006-0674-6833} \and
Luca Trautmann \orcidID{0009-0001-3517-5471} \and
Ivor Simpson \orcidID{0000-0001-5605-6626}}
\authorrunning{Pranav et al.}
\institute{LILI Lab, University of Sussex, Brighton, UK \\
\url{https://lililab-sussex.github.io/} \\
\email{\{P.Indrkanti, L.Trautmann, I.Simpson\}@sussex.ac.uk}}



\maketitle              
\begin{abstract}
We present an unsupervised single image bidirectional Magnetic Resonance Image (MRI) synthesizer that synthesizes an Ultra-Low Field (ULF) like image from a High-Field (HF) magnitude image and vice-versa.
Unlike existing MRI synthesis models, our approach is inspired by the physics that drives contrast changes between HF and ULF MRIs. Our forward model simulates a HF-to-ULF transformation by estimating the tissue-type Signal-to-Noise ratio (SNR) values based on target contrast values. For the Super-Resolution task, we used an Implicit Neural Representation (INR) network to synthesize HF image by simultaneously predicting tissue-type segmentations and image intensity without observed HF data. The proposed method is evaluated using synthetic ULF data generated from standard 3T T$_1$-weighted images for quantitative assessments and paired 3T-64mT T$_1$-weighted images for validation experiments. The proposed method achieved a 4x improvement in White Matter-Gray Matter contrast on synthetic ULF-like images and 2.5x improvement on 64mT images.
Sensitivity experiments demonstrated the robustness of our forward model to variations in target contrast, noise and initial seeding. Our contrast recovery experiment validates the reliability of enhanced contrast by comparing it with theoretical contrast.

\end{abstract}
\section{Introduction}
In Magnetic Resonance Imaging (MRI), field strengths, $B_0 \geq$ 1T are considered as clinical field strengths due to their ability to produce high quality, high resolution structural images. However, there exists a global disparity in the availability of MRI scanners, with accessibility to only one-tenth of the population \cite{mr-physics-low-field}. Ultra Low-Field (ULF) MRI scanners, ($B_0 < 0.1$) offer a promising alternative to High-Field (HF) scanners as they are accessible, cost-effective, sustainable and can also be portable \cite{LoHiResGAN}. However, ULF acquisitions lack the diagnostic quality of HF systems due to low Signal-to-Noise Ratio (SNR), low spatial resolution and poor tissue contrast, particularly affecting the Gray Matter-White Matter contrast \cite{textbook_mri,mr-physics-low-field,zhaolinUltraLowFieldMRIEnhancement2025}.
\par{
Cross-modality synthesis tools maximize scan utility and enable translation between imaging domains \cite{iglesiasQuantitativeBrainMorphometry2023,LoHiResGAN,StochDecimSim,lowfieldsim,zhaolinUltraLowFieldMRIEnhancement2025}. ULF-HF synthesis can bridge established clinical workflows with emerging ULF systems, accelerating clinical adoption of ULF acquisitions. Potential applications include longitudinal studies, calibration setups, tumor characterization, feasibility \& validation studies, and accessibility planning. One of the most promising applications of this synthesis framework is in longitudinal studies. Longitudinal neuroimaging studies require multiple acquisitions to capture morphological changes in the brain \cite{shuaibuCapturingLongitudinalChanges2025}. However, acquiring repeated HF scans is expensive and ULF systems could offer affordable follow-up scans. Our bidirectional synthesis framework facilitates seamless longitudinal registration across field strengths and spatial resolutions by enhancing tissue contrast.
}

\par{
While current state-of-the-art deep learning approaches \cite{dayarathnaUltraLowFieldHighField2024,super_res_Bouter,iglesiasQuantitativeBrainMorphometry2023,LoHiResGAN,lagunaSuperresolutionPortableLowfield2022} for ULF-HF image synthesis achieve notable performance, they are fundamentally opaque. This black-box nature, coupled with the known vulnerability to out-of-distribution (OOD) samples severely limits interpretability and hinders clinical adoption \cite{supervised_issues}. Additionally, these supervised models require large paired datasets which are expensive to acquire.}


\par{
To mitigate these issues, we propose an inspectable, physics-inspired, trustworthy single image bidirectional HF-ULF synthesis framework for magnitude images.
Our key contributions are:
\textit{(i)} A novel ULF synthesis framework with a forward model that estimates tissue-specific contrast factor vector, $\mathbf{m}$ by leveraging target ULF contrast characteristics;
\textit{(ii)} An unsupervised HF synthesis framework using Implicit Neural Representations (INRs) that integrates this forward model to jointly predict tissue-type segmentations and HF image.
We evaluate the robustness of our proposed approach by investigating the sensitivity of target contrast and random seed initialization on predicted HF images.
The code will be publicly available on
\textit{https://github.com/pranav-ind/hulfsynth}
.
}

\section{Related Works}
\label{sec:related}
Existing Low-Field simulators are broadly categorized into: (i) Naive (ii) Machine Learning (ML)-based and (iii) Bio-physical models. Naive synthesizers \cite{wangDeepLearningImage2021} employ a simple downsampling operation and add Gaussian noise without generally accounting for domain specificity like variation in SNR between HF and ULF. Data-driven models \cite{LoHiResGAN,LowGAN} trained on paired HF-ULF data learn direct domain mappings between ULF-HF, but are not trustworthy due to explainability and generalization issues \cite{supervised_issues}.
Biophysical models \cite{StochDecimSim,prob_deci_sim,lowfieldsim} capture the underlying physics of MR signal formation through field strength correction, relaxation time correction, noise modeling and SNR re-scaling. StocDeciSim \cite{StochDecimSim} modeled the inherent signal degradation of low-field systems by stochastically downsampling HF data and applying an empirical estimation for SNR rescaling. While this forward-modeling approach generates realistic, low-field-like synthetic images, its SNR rescaling method is strictly empirical and lacks generalizability to ULF settings. While our proposed ULF synthesis framework builds upon the foundational concept of SNR rescaling introduced by StocDeciSim \cite{StochDecimSim}, it extends current literature by replacing their strictly empirical approach with a modular target contrast approach.

\par{
Researchers have employed a wide array of architectural backbones to tackle ULF to HF synthesis, including Convolutional Neural Networks (CNN) \cite{SynthSRPublicAI,super_res_Bouter,iglesiasQuantitativeBrainMorphometry2023,lagunaSuperresolutionPortableLowfield2022}, Generative Adversarial Networks (GAN) \cite{LoHiResGAN}, Diffusion Models \cite{diffusion_IQT}, and INRs \cite{zhaolinUltraLowFieldMRIEnhancement2025}. Despite this architectural diversity, intrinsic trustworthiness remains a systemic challenge across existing methods. While most of these approaches formulate the synthesis task as an inverse problem, they rely heavily on paired datasets and solve in a supervised fashion. Consequently, their embedded forward models fail to explicitly capture the biophysical principles that actually govern signal intensity transitions between HF and ULF regimes. SynthSeg \cite{synthseg} is a robust CNN-based segmentation algorithm capable of handling diverse MRI contrasts including low-field acquisitions, without retraining. Given the extensive diversity of its training data across field strengths and contrast types, we choose SynthSeg as our segmentation baseline for evaluations.
}

\par{
INRs are continuous and differentiable signal representations, parameterized by neural networks.
Recent advances in INRs, particularly sinusoidal representation networks \cite{sitzmannImplicitNeuralRepresentations2020} (SIRENs) and Wavelet Implicit Neural Representations \cite{saragadamWIREWaveletImplicit2023} (WIREs) have enabled effective modeling of images along with their derivatives.
INRs are actively used for a wide range of medical imaging tasks including Image Reconstruction \cite{wuIREMHighResolutionMagnetic2021a}, Registration \cite{shuaibuCapturingLongitudinalChanges2025}, Style Transfer \cite{zhaolinUltraLowFieldMRIEnhancement2025}, Segmentation and Compression \cite{molaeiImplicitNeuralRepresentation2023}.
INRs are well-suited for Super-Resolution tasks as they are resolution-agnostic, memory-efficient and can be fitted to a single volume, thereby mitigating auxiliary data bias \cite{molaeiImplicitNeuralRepresentation2023,yeSuperResolutionBiomedicalImaging2023}. 
}

\section{Methods}
\label{sec:methods}
Our proposed bidirectional synthesizer, HULFSynth, is divided into two components: 
(i) Ultra Low-Field synthesis (ULFSynth): generates a synthetic ULF image given a HF image
(ii) High-Field synthesis (HFSynth): super-resolves a HF-like image given an ULF image
\subsection{Ultra Low-Field Synthesis}
Given, a 3D HF magnitude image and its tissue-type segmentations $\mathrm{\{X, S_t\}}$, we generate a synthetic ULF-like image, Y by estimating a tissue-type contrast factor vector, $\mathbf{m}$. We divide our ULFSynth pipeline into: (i) {Contrast factor estimation} and (ii) {Contrast modulation}.
$\mathbf{m}$ quantifies the relative change in signal intensity for each tissue-type between HF and ULF. Applying $\mathbf{m}$, we perform Contrast Modulation to generate a synthetic ULF-like image.

\textbf{{Contrast factor estimation:}}
We map HF image to segmentations for tissue-type intensities, $\mathrm{img_t}$,
where  t $\mathrm{\in}$ T = $\{$WM, GM, CSF$\}$; WM = White Matter, GM = Gray Matter, CSF = Cerebrospinal fluid.
We estimate $\mathrm{\mathbf{m} \in \mathbb{R}^{3}}$ from tissue-type SNRs, $\mathbf{A} \in \mathbb{R}^{3 \times 3}$ and target ULF contrast, $\mathrm{\mathbf{c} \in \mathbb{R}^{3}}$. SNRs are computed from manually selected flat regions of interest (ROIs) as the ratio of mean tissue intensity to standard deviation of background noise, $\sigma_\mathrm{bg}$ in accordance with National Electrical Manufacturers Association (NEMA) standards (Eq. \ref{eq:snr_t}). A Rayleigh correction factor of 1.53 is applied to account for the Rayleigh distribution of background noise in magnitude images \cite{mri_book}.
\begin{equation}
\label{eq:snr_t}
    \mathrm{SNR_t = {\frac{\mu_{ROI_{img_{t}}}}{\sigma_{bg} \times 1.53}}}  
\end{equation}

\par{
Target contrast, $\mathbf{c}$ is empirically computed by calculating the difference of SNRs between two tissues of ULF images. 
In T$_1$W images, it is known apriori that the signal intensity of WM is greater than GM and CSF \cite{mri_book}. We incorporate this prior known MR physics relationship between tissues types in T$_1$W to formulate $\mathbf{A}$. From the element-wise relationship, $\mathrm{c_{ij} = m_{i} S_{i} - m_{j} S_{j}}$, a system of contrast equations, with 3 equations and 3 unknowns is constructed (Eq. \ref{eq:contrast_system}), where  i, j $\mathrm{\in T}$
and $\mathrm{c_{wc}, c_{wg}, c_{gc}}$ are WM-CSF, WM-GM and GM-CSF contrasts respectively.
\begin{equation}
\label{eq:contrast_system}
\underbrace{
\begin{bmatrix}
\mathrm{SNR_{WM}} & 0 & \mathrm{-SNR_{CSF}} \\
\mathrm{SNR_{WM}} & \mathrm{-SNR_{GM}} & 0\\
0 & \mathrm{SNR_{GM}} & \mathrm{-SNR_{CSF}} \\
\end{bmatrix} }_\mathbf{A}       
\underbrace{
 \begin{bmatrix}
\mathrm{m_{WM}} \\
\mathrm{m_{GM}} \\
\mathrm{m_{CSF}} \\
\end{bmatrix} }_\text{$\mathbf{m}$}  = 
\underbrace{
\begin{bmatrix}
\mathrm{c_{wc}} \\
\mathrm{c_{wg}} \\
\mathrm{c_{gc}} \\
\end{bmatrix}  }_\text{$\mathbf{c}$}
\end{equation}

\begin{equation} \label{eq:optim_prob}
    \mathrm{\underset{0 \leq m_t \leq 1}{\text{min}} \: \: \frac{1}{2} || \mathbf{A{m} - {c}} ||_{2}^{2}  + \epsilon \: || {\mathbf{m}} || ^{2}}
\end{equation}
where, $\epsilon$ is regularization strength. Eq. \ref{eq:optim_prob} is a bounded Least Squares optimization problem, which is solved with grid search.

\par{
\textbf{Contrast Modulation}: 
 The mapped tissue-type intensities are smoothed with Gaussian kernel (G$_\sigma$) and downsampled ($\mathrm{\downarrow_{df}}$). The resultant tissue-type intensities are degraded using $\mathbf{m}$ to simulate ULF contrast and further corrupted with Rician noise, $\gamma_{\{\rho,\sigma_r\}}$. These modulated tissue-type intensities are recombined to yield a synthetic ULF-like image that accurately reflects key ULF properties: low contrast-to-noise ratio (CNR), reduced spatial resolution, increased blurring and noise. Our \textbf{forward model}, $\phi$ is defined as, 
\begin{equation}
\label{eq:ulfsynth}
\mathrm{Y = \phi(S_t, X, m_t, G_\sigma, \gamma_{\{\rho,\sigma_r\}}) =  \sum_{t \in T}^{} \downarrow_{df} ((B(X) \odot S_t) * G_{{\sigma}}) \: m_t + \gamma_{\{\rho,\sigma_r\}}}
\end{equation}
}

\subsection{High-Field Synthesis: Super-Resolution}
\begin{figure*}[htbp]
        \includegraphics[width=1.1\textwidth]{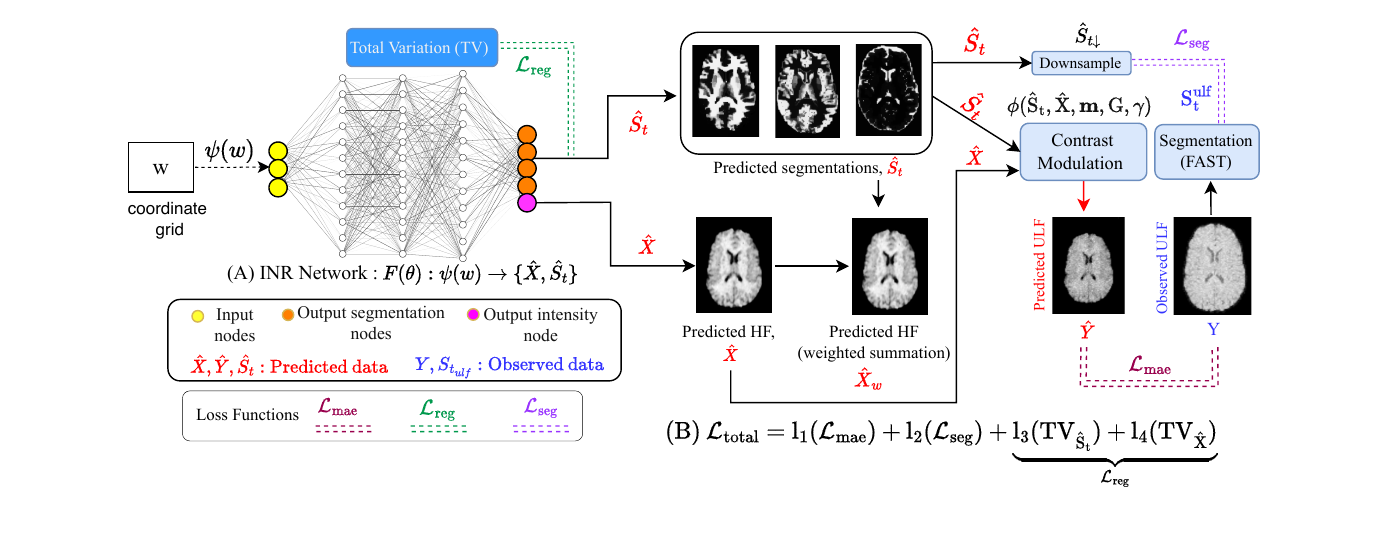}
    \caption{High-Field Synthesis (HFSynth) pipeline: (A) Given an observed ULF image and its segmentations (blue variables), our model jointly predicts HF-like image intensity and soft tissue segmentations (red variables) without the need for HF supervision. This is achieved by formulating our forward model ($\phi$) within the INR framework. At inference, INR predicts HF-like images at arbitrary resolutions.
    (B) Loss functions that govern the learning process, where $\mathrm{l_1, l_2, l_3, l_4}$ are tunable hyperparameters.
    }
    \label{fig:block_diagram_2}
    
\end{figure*}

Given a 3D ULF image and its tissue segmentations, $\mathrm{\{Y, S_t^{ulf}}\}$, we jointly predict the latent HF-like image and its segmentations without any HF supervision, as depicted in Fig. \ref{fig:block_diagram_2}(A). To solve this ill-posed, inverse problem, we parameterize an INR using a Multilayer perceptron (MLP) with normalized 3D coordinates, w = (x, y, z) as input, whose coordinates are in the range $\Omega \in$ [-1, 1].
Fourier Feature Embedding, $\psi(w)$ was applied to input coordinates to enable MLP learn high-frequency functions \cite{molaeiImplicitNeuralRepresentation2023}.
The INR, $F(.)$ parametrizes images as continuous functions where embedded spatial coordinates are mapped to HF intensity, $\mathrm{\hat{X}}$ and tissue-type segmentations, $\mathrm{\hat{S}_t}$ values such that $F_\theta \mathrm{: \psi(w) \rightarrow \{\hat{X} , \hat{S_t}\}}$, where $\theta$ represents MLP parameters.
The final predicted HF image, $\mathrm{\hat{X}_{w}}$ is reconstructed as a weighted sum of predicted tissue segmentations and intensity i.e., $\mathrm{ \hat{X}_{w} = \underset{t \in T}\Sigma \hat{X} \odot \hat{S}_t }$.

\par{
\textbf{Network: }
Adapted from WIRE \cite{saragadamWIREWaveletImplicit2023}, we implemented an INR network using an MLP with Gabor wavelet activations. The selection of the WIRE activation function is motivated by the need for a specific inductive bias: one that demonstrates high fidelity in reconstructing high-frequency details while exhibiting low sensitivity to noise. Additionally, unlike SIRENs \cite{sitzmannImplicitNeuralRepresentations2020}, WIRE does not require carefully initialized weights.
The network configuration is, input spatial features: 3, output features: 5, hidden layers: 3, features per layer: 128, optimizer: Adam. In the final layer, segmentation outputs (4) were passed to softmax function and image intensity output (1) was passed to a ReLU function. Pre-activation outputs were used for regularization. 
}

\textbf{Loss Functions: }
The optimization problem depicted in Fig. \ref{fig:block_diagram_2}(B) minimizes a loss function consisting of reconstruction, segmentation and regularization terms. To capture voxel-wise differences, {Mean Absolute Error} $\mathcal{L}_{\mathrm{mae}}$ was used. A fusion of Dice loss and Cross Entropy loss (Eq. 22 in Ref. \cite{maLossOdysseyMedical2021}) was used for segmentation prediction $(\mathcal{L}_{\mathrm{seg}})$ to promote gradient stability and handle class-imbalances. Total Variation (TV) prior was used to promote linear piece-wise smoothness in reconstructed images while preserving edges.

\section{Experiments and Results}
\label{sec:experiments}
\textbf{Datasets and Implementation: }
The proposed method was evaluated on two publicly available datasets: \textit{(i)} IXI Dataset \cite{IXIDatasetBrain}, consisting of HF (3T) T$_1$-weighted images with spatial dimensions: $150 \times 256$ and 256 slices per volume. \textit{(ii)} LMIC Dataset \cite{vandenbroekPaired64mT3T2025a} comprising paired HF-ULF (3T-64mT) T$_1$-weighted images with spatial dimensions: $112 \times 136$ and 40 slices per volume.
Observed HF and ULF images were preprocessed using BET \cite{smithFastRobustAutomated2002} and FAST \cite{zhangSegmentationBrainMR2001} with bias field correction.
HF images were affine-registered to ULF space for evaluations (Experiment-2). Intensity images were normalized to [0,1].
INR was trained with voxel patches using PyTorch Lightning on an NVIDIA RTX A6000 GPU (48GB). The average run time of IXI and LMIC datasets were 17 and 11 minutes per volume respectively. The average memory usage of IXI and LMIC datasets were 39 and 6 Gb per volume respectively.
\par{
\textbf{Image Assessment: } Image Quality was quantified using a combination of structural, luminance, image contrast and cohesion metrics: Structural Similarity Index (SSIM) and Mean Shifted Line Correlation (MSLC) \cite{dohmenSimilarityQualityMetrics2025}. To measure perceptual fidelity, we used Learned Perceptual Image Patch Similarity (LPIPS) with an Alex-net backbone in the axial plane \cite{dohmenSimilarityQualityMetrics2025}. 
Segmentation predictions were quantified using Dice score and Intersection over Union (IoU).
We approximated the standard deviation of background noise with CSF signal to quantify WM-GM contrast enhancement in the predicted images.
}

\par{
\textbf{Hyperparameter Tuning: }Lagrangian multipliers (in Fig. \ref{fig:block_diagram_2}(B))
were carefully tuned with a fusion metric, Reconstruction Quality Score (RQS) which was composed using ULF prediction scores: SSIM, MSLC, LPIPS and ULF segmentation prediction scores Dice and IoU to find the default hyperparameter configuration with grid search. The default hyperparameter configuration was used for all experiments, given its stable convergence across synthetic and 64mT data, demonstrating strong hyperparameter transferability across datasets.
Default hyperparameter configuration is as follows: l$_1$: 1e3, l$_2$: 1e2, l$_3$: 9.5e-1, l$_4$: 1.5, Fourier Frequency: 72, Fourier scale: 4.0, epochs = 15e3 
}

\par{
\textbf{Baselines: }
We chose StocDeciSim \cite{StochDecimSim} as the baseline for the ULFSynth task because its SNR rescaling approach provides the most direct comparison to our proposed framework.
We compared our HFSynth method against Bicubic and Voxel Grid (trilinear) interpolations as they preserve observed data integrity.
We choose two state-of-the-art HF synthesis methods: LoHiResGAN \cite{LoHiResGAN}, which was trained on paired 64mT-3T data and SynthSR \cite{SynthSRPublicAI}, which was trained on real and synthetic low-field data. These baselines were selected because our experimental datasets likely have representations in their training data, enabling a fairer comparison without requiring additional retraining.
We used skull-stripped and bias field corrected ULF images as inputs to the LoHiResGAN and SynthSR models. SynthSeg \cite{synthseg} was used as a baseline for the comparison of predicted HF segmentations.
}

\subsection{Experiment 1: Qualitative Evaluation}
To evaluate the performance of both of ULF and HF synthesis methods, we randomly selected a cohort of 25 subjects (3T) from IXI dataset \cite{IXIDatasetBrain}. \\
\textbf{ULFSynth: }
HF magnitude images and their FAST segmentations were used to generate synthetic ULF images using ULFSynth pipeline.
Subject-specific tissue-type $\mathrm{SNR^{HF}}$ values were computed on HF images to estimate $\mathbf{m}$ (Eq. \ref{eq:optim_prob}) with target contrast set to $\mathbf{c}$ = [2, 12, 17], $\mathrm{\downarrow_{df}}$ as 2 and $\gamma_{\{\rho,\sigma_r\}} = \{5, 15\} $.
The range of computed SNR values for WM, GM and CSF were 58 to 156, 48 to 116 and 21 to 55 respectively. The search space of initial $\mathbf{m}$ and $\epsilon$ ranged from 0.0 to 0.5.
StocDeciSim was implemented as a comparative baseline using its default parameters: $\mathrm{SNR_{WM}} = 64.50$, $\mathrm{SNR_{GM}} = 54.14$, and a decimation factor of 2. Qualitative results from IXI dataset are illustrated in Fig. \ref{fig:ulfsynth_} to visually assess the impact of tissue contrast, blurriness, and noise on the synthetic outputs.
\begin{figure}[htbp]

  \centering
  \centerline{\includegraphics[width=\linewidth]{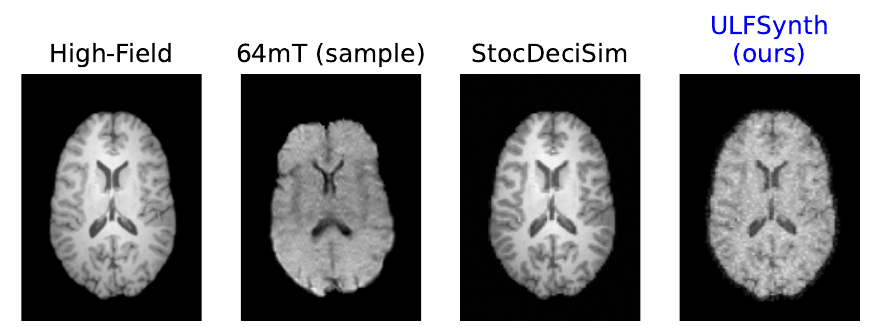}}

\caption{Qualitative Comparison of ULFSynth in Axial plane. The columns (left-right) represent: (a) Reference High-Field (IXI dataset) (b) a representative 64mT acquisition from the LMIC dataset, (c) the StocDeciSim baseline (with Gaussian noise), (d) Our proposed \textit{ULFSynth} (with Rician noise). Note that column (b) displays an unpaired subject, included solely as a representative ULF sample for visual aid. Qualitatively, our method generates more accurate Ultra Low-Field-like tissue contrast with realistic noise texture compared to the baseline.}
\label{fig:ulfsynth_}
\end{figure}

\textbf{HFSynth: }
Using synthetic ULF image and its segmentations (estimated from FAST) as observed data, we train the INR to jointly predict HF image and segmentations. Quantitative results of the reconstructed image quality are summarized in Table ~\ref{tab:Quantiative}, and qualitative results are visualized in Fig. \ref{fig:qualitative}.
Our method achieved good edge preservation, reduced noise and enhanced WM-GM contrast ($\mathrm{c_{wg}}$ = 7.31), compared to the baselines. Mean Dice and IoU scores were: 0.6 and 0.46 respectively. WM-GM contrast values for LoHiResGAN and SynthSR on the IXI dataset were not computed, as SynthSeg failed to produce reliable tissue segmentations required for contrast quantification.

\begin{table}[htbp]
\centering
\caption{Quantitative comparison of the reconstruction performance of our proposed HFSynth approach against different baselines on IXI and LMIC datasets. HFSynth yielded a 4x WM-GM contrast improvement on synthetic ULF data (IXI) and a 2.5x improvement on 64mT data (LMIC) against a 4\% tradeoff on image quality metrics. All metrics are reported as mean (stdev). N.A. = Not Available. }
\begin{tabular}{|c|cc|cc|cc|cc|}
\hline
 & \multicolumn{2}{c|}{\textbf{SSIM} $\uparrow$} & \multicolumn{2}{c|}{\textbf{LPIPS} $\downarrow$} & \multicolumn{2}{c|}{\textbf{MSLC} $\downarrow$} & \multicolumn{2}{c|}{\textbf{\begin{tabular}[c]{@{}c@{}}WM-GM\\ contrast $\uparrow$\end{tabular}}} \\ \hline
 & \multicolumn{1}{c|}{IXI} & LMIC & \multicolumn{1}{c|}{IXI} & LMIC & \multicolumn{1}{c|}{IXI} & LMIC & \multicolumn{1}{c|}{IXI} & LMIC \\ \hline
Bicubic & \multicolumn{1}{c|}{\begin{tabular}[c]{@{}c@{}}0.67\\ (0.05)\end{tabular}} & \begin{tabular}[c]{@{}c@{}}0.80\\ (0.04)\end{tabular} & \multicolumn{1}{c|}{\begin{tabular}[c]{@{}c@{}}0.61\\ (0.02)\end{tabular}} & \begin{tabular}[c]{@{}c@{}}0.14\\ (0.02)\end{tabular} & \multicolumn{1}{c|}{\begin{tabular}[c]{@{}c@{}}0.37\\ (0.05)\end{tabular}} & \begin{tabular}[c]{@{}c@{}}0.25\\ (0.05)\end{tabular} & \multicolumn{1}{c|}{\begin{tabular}[c]{@{}c@{}}1.49\\ (0.69)\end{tabular}} & \begin{tabular}[c]{@{}c@{}}1.59\\ (0.16)\end{tabular} \\ \hline
\begin{tabular}[c]{@{}c@{}}VoxelGrid\\ (trilinear)\end{tabular} & \multicolumn{1}{c|}{\textbf{\begin{tabular}[c]{@{}c@{}}0.92\\ (0.02)\end{tabular}}} & \textbf{\begin{tabular}[c]{@{}c@{}}0.81\\ (0.04)\end{tabular}} & \multicolumn{1}{c|}{\begin{tabular}[c]{@{}c@{}}0.14\\ (0.03)\end{tabular}} & \textbf{\begin{tabular}[c]{@{}c@{}}0.14\\ (0.02)\end{tabular}} & \multicolumn{1}{c|}{\begin{tabular}[c]{@{}c@{}}0.35\\ (0.05)\end{tabular}} & \begin{tabular}[c]{@{}c@{}}0.25\\ (0.05)\end{tabular} & \multicolumn{1}{c|}{\begin{tabular}[c]{@{}c@{}}1.78\\ (0.89)\end{tabular}} & \begin{tabular}[c]{@{}c@{}}1.59\\ (0.16)\end{tabular} \\ \hline
LoHiResGAN & \multicolumn{1}{c|}{N.A.} & \begin{tabular}[c]{@{}c@{}}0.16\\ (0.03)\end{tabular} & \multicolumn{1}{c|}{N.A.} & \begin{tabular}[c]{@{}c@{}}0.21\\ (0.03)\end{tabular} & \multicolumn{1}{c|}{N.A.} & \begin{tabular}[c]{@{}c@{}}0.23\\ (0.03)\end{tabular} & \multicolumn{1}{c|}{N.A.} & \begin{tabular}[c]{@{}c@{}}1.16\\ (0.43)\end{tabular} \\ \hline
SynthSR & \multicolumn{1}{c|}{\begin{tabular}[c]{@{}c@{}}0.1\\ (0.008)\end{tabular}} & \begin{tabular}[c]{@{}c@{}}0.1\\ (0.01)\end{tabular} & \multicolumn{1}{c|}{\begin{tabular}[c]{@{}c@{}}0.48\\ (0.1)\end{tabular}} & \begin{tabular}[c]{@{}c@{}}0.36\\ (0.05)\end{tabular} & \multicolumn{1}{c|}{\begin{tabular}[c]{@{}c@{}}0.57\\ (0.08)\end{tabular}} & \begin{tabular}[c]{@{}c@{}}0.24\\ (0.2)\end{tabular} & \multicolumn{1}{c|}{N.A.} & \begin{tabular}[c]{@{}c@{}}2.73\\ (0.95)\end{tabular} \\ \hline
\textbf{\begin{tabular}[c]{@{}c@{}} HFSynth \\ (ours) \end{tabular}} & \multicolumn{1}{c|}{\begin{tabular}[c]{@{}c@{}}0.88\\ (0.03)\end{tabular}} & \begin{tabular}[c]{@{}c@{}}0.77\\ (0.06)\end{tabular} & \multicolumn{1}{c|}{\textbf{\begin{tabular}[c]{@{}c@{}}0.14\\ (0.04)\end{tabular}}} & \begin{tabular}[c]{@{}c@{}}0.16\\ (0.04)\end{tabular} & \multicolumn{1}{c|}{\textbf{\begin{tabular}[c]{@{}c@{}}0.34\\ (0.05)\end{tabular}}} & \textbf{\begin{tabular}[c]{@{}c@{}}0.23\\ (0.06)\end{tabular}} & \multicolumn{1}{c|}{\textbf{\begin{tabular}[c]{@{}c@{}}7.31\\ (3.68)\end{tabular}}} & \textbf{\begin{tabular}[c]{@{}c@{}}3.96\\ (1.91)\end{tabular}} \\ \hline
\end{tabular}
\label{tab:Quantiative}
\end{table}

\begin{figure}[]

  \centering
  \centerline{\includegraphics[width=\linewidth]{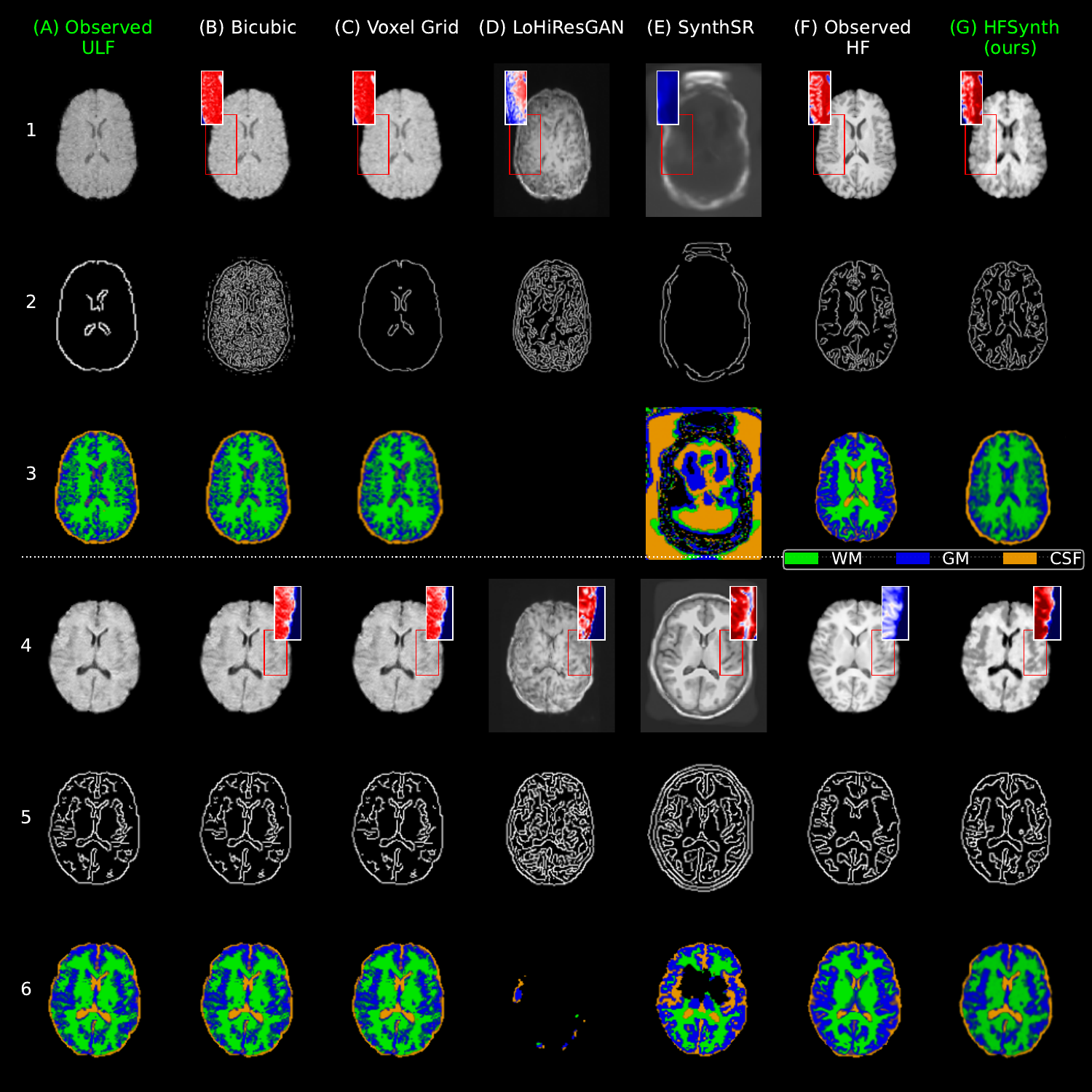}}

\caption{Qualitative comparison of HF Synthesis (Axial plane).
Rows 1 and 4 show subjects from IXI and LMIC datasets, with corresponding Canny edge maps in Rows 2 and 5. Subplots A(1), A(4) are observed ULF inputs and A(3), A(6) are FAST segmentations of ULF inputs. SynthSeg method was used to segment the predictions of baselines (LoHiResGAN and SynthSR). While the interpolation baselines (Bicubic and VoxelGrid) preserve data integrity, they failed to enhance tissue contrast. Supervised approaches, LoHiResGAN and SynthSR failed on synthetic ULF data (subplots D(1) and E(1)), demonstrating their vulnerability to OOD samples. Compared to the baselines (columns B-E), our proposed HFSynth method (column G) demonstrates enhanced WM-GM contrast, reduced noise and sharper edges.
Quantitative results are presented in Table~\ref{tab:Quantiative}.
}
\label{fig:qualitative}
\end{figure}

\subsection{Experiment 2: Validation on 64mT data}
To assess the validity of the HFSynth method on real ULF (64mT) data, we used the available paired ULF-HF subjects ($\mathrm{N_{samples} = 10}$) from the LMIC dataset \cite{vandenbroekPaired64mT3T2025a}. The paired HF volumes were only used for evaluation, not for training the INR.
$\mathrm{SNR_{t}^{HF}}$ values were chosen arbitrarily from IXI dataset. 
Subject-specific target contrast factor vector, $\mathbf{c}$ (E.g., $\mathbf{c}$ = [11, 37, 26]) was computed for each ULF subject to estimate $\mathbf{m}$. These computed contrast vectors were
used as targets to estimate the latent $\mathbf{m}$ in Eq. \ref{eq:ulfsynth}. 
Predicted HF images showed enhanced WM-GM contrast (3.96) with preserved structural fidelity and smoother segmentations compared to baselines (See Fig. \ref{fig:qualitative}, Table \ref{tab:Quantiative}). HF segmentations predicted by HFSynth yielded higher mean evaluation metrics (Dice: 0.67, IoU: 0.50) than those predicted by the SynthSeg baseline (Dice: 0.33, IoU: 0.25).

\subsection{Experiment 3: Sensitivity Analysis}
\begin{figure}[htb]

\centering
{\includegraphics[width=0.9\linewidth]{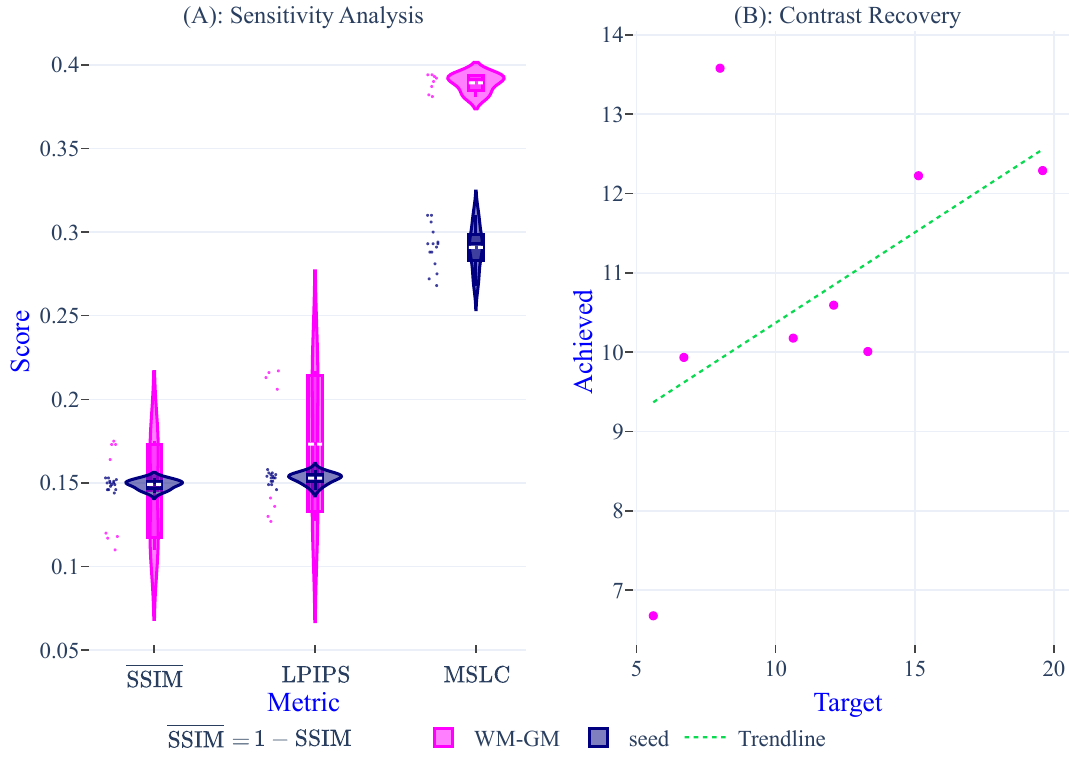}}


\caption{(A) Exploring sensitivity of target contrasts and random seed initializations. Low variance of image metrics ($\sigma^2 =$ 1e-3, 2e-3 and 1e-4 for SSIM, LPIPS and MSLC respectively) across range of target contrasts indicates that our framework is generalizable to a wide range of plausible ULF contrast settings. Low sensitivity to variations of the initialization ($\sigma^2 =$ 1e-5, 1e-5 and 1.7e-4 for SSIM, LPIPS and MSLC respectively) demonstrates model's robustness.
(B) Contrast Recovery: Achieved vs target WM-GM contrast with Least Squares fit ($\mathrm{R^2=0.26}$) suggests a positive trend in contrast fidelity in reconstructed images for a limited range of target contrasts.
}
\label{fig:sensitivity}
\end{figure}

\textbf{Random Seed Initialization: }
To investigate the stochastic effects of initialization on our INR predictions, we conducted a sensitivity assessment.
One LMIC subject (randomly selected) was used to train the model with 3 distinct random seeds and 5 independent runs per seed. The variance of image quality metrics was negligible (SSIM: 1e-5, LPIPS: 1e-5, MSLC: 7e-4), suggesting that our HFSynth approach is robust to random seed initialization.

\textbf{Target Contrast: }
In both the ULFSynth and HFSynth pipelines, the parameter $\mathbf{c}$ governs the estimation of $\mathbf{m}$, which subsequently drives the effective contrast modulation. To investigate the robustness of our framework against contrast variability, we conducted a sensitivity analysis using a range of plausible values of $\mathbf{c}$. 

\par{
To evaluate the model's sensitivity to plausible variations in $\mathbf{c}$, we randomly selected a single IXI subject (3T) to generate a cohort of 8 synthetic ULF volumes. These volumes were synthesized with varying target contrast vectors to cover a broad range: $\mathrm{c_{wg}}$ from 5 to 20, $\mathrm{c_{wc}}$ from 20 to 75 and $\mathrm{c_{gc}}$ from 15 to 55. These values for target $\mathbf{c}$ were computed empirically on 64mT data (LMIC dataset). The reconstruction quality of predicted HF volumes remained stable across all sensitivity samples. The variance in evaluation metrics was $\leq 9\%$ (SSIM: 1e-3, LPIPS: 2e-3, MSLC: 1e-4), suggesting that our framework is robust and adaptable to a wide range of target contrasts. See Fig. \ref{fig:sensitivity}(A).
}

\par{
\textbf{Contrast Recovery: }
To quantify the effective contrast recovery, we analyzed the linearity of the contrast (achieved $\mathrm{c_{wg}}$ vs target $\mathrm{c_{g}}$) by modeling the relationship between the target input contrast and the achieved output contrast. An Ordinary Least Squares (OLS) fit applied to the $\mathrm{c_{wg}}$ metric resulted in a modest $\mathrm{R^2}$ of $0.26$ (Adjusted $\mathrm{R^2} = 0.135$), exhibiting a positive monotonic trend. As illustrated in Fig. \ref{fig:sensitivity}(B), our Contrast Recovery experiment quantifies the limited interpretable contrast fidelity in reconstructed images for the given range of target contrasts.
}

\subsection{Ablation Studies}
An ablation study was conducted to assess the role of each loss function component. As detailed in Table \ref{tab:ablation}, we trained the model by systematically excluding each component i.e., reconstruction term, $\mathrm{l_1}=0$, segmentation term, $\mathrm{l_2}=0$, and regularization terms, $\mathrm{l_3, l_4}=0$.
This study affirms that all the loss function components contribute positively to the joint optimization framework. These models were trained on LMIC dataset.

\textbf{Oracle baseline: }
Applying the FSL FAST \cite{zhangSegmentationBrainMR2001} algorithm to noisy ULF (64mT) data introduces estimation errors that can propagate through our HFSynth pipeline. To isolate this confounding effect, we conducted an "Oracle Baseline" that substitutes ULF segmentations with proxy gold-standard estimates from corresponding paired HF acquisitions. This "ideal scenario" outperformed the default setup (SSIM: 6\%, LPIPS: 4\%, Dice: 12\%, IoU: 20 \%. See Table \ref{tab:ablation}), confirming that suboptimal ULF segmentation is a critical bottleneck in the proposed HFSynth method.

\begin{table}[htbp]
\centering
\caption{ Ablation study evaluating individual loss components. Image (SSIM, LPIPS) and Segmentation (Dice, IoU) evaluation metrics are compared across configurations with default hyperparameters. The 'Oracle Baseline' substitutes ULF segmentations with High-Field segmentations (proxy for gold-standard). Metrics are reported as mean (stdev).}
\begin{tabular}{@{}ccccc@{}}
\toprule
\textbf{Component} &
  \textbf{SSIM $\uparrow$} &
  \textbf{LPIPS $\downarrow$} &
  \textbf{Dice $\uparrow$} &
  \textbf{IoU $\uparrow$} \\ \midrule
$\mathrm{l_1}$ = 0 &
  \begin{tabular}[c]{@{}c@{}}0.56 \\ (0.07)       \end{tabular} &
  \begin{tabular}[c]{@{}c@{}}0.48 \\ (0.06)       \end{tabular} &
  \begin{tabular}[c]{@{}c@{}}0.31 \\ (0.01)        \end{tabular} &
  \begin{tabular}[c]{@{}c@{}}0.5  \\ (0.02)        \end{tabular} \\
$\mathrm{l_2}$ = 0 &
  \begin{tabular}[c]{@{}c@{}}0.068 \\(0.01) \end{tabular} &
  \begin{tabular}[c]{@{}c@{}}0.52 \\(0.04) \end{tabular} &
  \begin{tabular}[c]{@{}c@{}}0.3 \\(0) \end{tabular} &
  \begin{tabular}[c]{@{}c@{}}0.1 \\(0) \end{tabular} \\
$\mathrm{l_3}$ = 0 &
  \begin{tabular}[c]{@{}c@{}}0.76 \\ (0.06)\end{tabular} &
  \begin{tabular}[c]{@{}c@{}}0.17 \\ (0.03)\end{tabular} &
  \begin{tabular}[c]{@{}c@{}}0.57 \\ (0.01)\end{tabular} &
  \begin{tabular}[c]{@{}c@{}}{0.49} \\ (0.01)\end{tabular} \\
$\mathrm{l_4}$ = 0 &
  \begin{tabular}[c]{@{}c@{}}0.75 \\ (0.06)\end{tabular} &
  \begin{tabular}[c]{@{}c@{}}0.16 \\ (0.02)\end{tabular} &
  \begin{tabular}[c]{@{}c@{}}0.61 \\ (0.01)\end{tabular} &
  \begin{tabular}[c]{@{}c@{}}0.51 \\ (0.02)\end{tabular} \\
ORACLE BASELINE &
  \begin{tabular}[c]{@{}c@{}}\textbf{0.83} \\ (0.04)\end{tabular} &
  \begin{tabular}[c]{@{}c@{}}\textbf{0.12} \\ (0.01)\end{tabular} &
  \begin{tabular}[c]{@{}c@{}}\textbf{0.79} \\ (0.03)\end{tabular} &
  \begin{tabular}[c]{@{}c@{}}\textbf{0.7} \\ (0.01)\end{tabular} \\
\begin{tabular}[c]{@{}c@{}}$l_1 + l_2 + l_3 + l_4$ \end{tabular} &
  \begin{tabular}[c]{@{}c@{}}0.77 \\ (0.06)\end{tabular} &
  \begin{tabular}[c]{@{}c@{}}0.16 \\ (0.04)\end{tabular} &
  \begin{tabular}[c]{@{}c@{}}0.67 \\ (0.01)\end{tabular} &
  \begin{tabular}[c]{@{}c@{}}0.50 \\ (0.01)\end{tabular} \\ \bottomrule
\end{tabular}

\label{tab:ablation}
\end{table}

\section{Discussion}
\label{sec:discussion}
We present a qualitative and quantitative evaluation of HFSynth framework on both synthetic and real (64mT) ULF data. As the forward model functions as an integrated component of the HFSynth pipeline, its quantitative evaluation is inherently coupled with HFSynth performance.
Through our sensitivity experiments, we validate the robustness of our HFSynth framework (See Fig. \ref{fig:sensitivity}A). Our contrast recovery experiment validates the contrast fidelity in reconstructed HF-like image by comparing achieved contrast in predicted HF images with target contrast (See Fig. \ref{fig:sensitivity}B). Qualitative results of ULFSynth pipeline are visualized in Fig. \ref{fig:ulfsynth_}. As the IXI datasets contains only HF (3T) volumes without paired ULF acquisitions, ULFSynth could not be quantitatively evaluated against its corresponding ULF reference.

\par{
\textbf{SSIM-Contrast trade-off: }
Although our proposed method significantly improves WM-GM contrast in the predicted HF images, it yields slightly lower SSIM ($\leq 4\%$) and LPIPS ($\leq 2\%$) scores compared to a simple VoxelGrid fit as shown in Table \ref{tab:Quantiative}. This tradeoff between standard reconstruction metrics and tissue contrast arises from the inherent conflict between strict voxel-wise fidelity and perceptual realism, formalized in the literature as the perception-distortion tradeoff \cite{blauPerceptionDistortionTradeoff2018}. Since standard reference metrics heavily penalize spatial and intensity deviations, they inadvertently suppress high-frequency structural details like sharpened tissue boundaries. By explicitly prioritizing tissue contrast enhancement, our model  potentially suffers an inherent tradeoff in SSIM score; crucially, this quantitative reduction reflects the metrics' sensitivity to intensity shifts rather than actual structural distortion.
Subsequently, we chose qualitative evaluation using edge maps (See Fig. \ref{fig:qualitative}) over absolute error maps to prioritize the precise delineation of tissue boundaries as absolute error maps are often dominated by low-frequency intensity shifts, which aggregate voxel-wise intensity errors. 
}

\par{
\textbf{Baseline performance analysis: } 
While our model performed better than LoHiResGAN, we note that the 64mT inputs were preprocessed slightly differently without 3T registration, limiting direct comparability with their original results \cite{LoHiResGAN}. While SynthSR \cite{SynthSRPublicAI} performed well on some subjects (See Fig. \ref{fig:qualitative}E(4)), it demonstrated inconsistent performance across the cohort, resulting in suboptimal quantitative metrics. This inconsistency likely reflects the presence of OOD characteristics in our evaluation data- both synthetic and real ULF images; that were not represented in the model's training distribution. Another critical limitation of both these approaches is hallucination. For instance, when tested with skull-stripped 64mT data, both models hallucinated skull structure in their predictions as they were trained on skull-intact data (See Fig. \ref{fig:qualitative}D(4), E(4)). These observations underscore a crucial trustworthiness issue in all the data-driven models. Recently, ULFINREnc \cite{zhaolinUltraLowFieldMRIEnhancement2025} presented an INR-based ULF enhancement framework using 7T data. Although a promising approach, it relies on external, unpaired data to transfer style characteristics onto 64mT acquisitions. Consequently, this method was excluded from our baseline comparisons as it does not offer interpretability and needs subject-specific retraining. 
Notably, our mechanistic approach ensures a trustworthy, hallucination-free synthesis as it relies only on the observed ULF data.
}

\par{
\textbf{Segmentation bottleneck: }
Despite our model's robustness to variations in contrast, noise and random initialization (Fig. \ref{fig:sensitivity}(A)), its overall performance is reliant on the performance of segmentation algorithm applied to ULF data. This presents a critical bottleneck, as classical segmentation tools designed specifically for HF data (E.g. FAST \cite{zhangSegmentationBrainMR2001}) may not perform consistently on noisy, low-contrast ULF data. We attribute the performance gap between synthetic ULF-like data and real 64mT data to this segmentation bottleneck.
The Oracle Baseline experiment quantitatively validates this hypothesis, by outperforming the baselines. (See Table \ref{tab:ablation}). 
}

\section{Conclusion}
\textbf{Summary: }
We introduce a modular, physics-driven bidirectional synthesis framework that jointly enhances tissue contrast, preserves structural detail, and suppresses noise without relying on external data or introducing hallucinated artifacts (style, noise or contrast).
By explicitly modeling contrast information within the INR framework, we achieve a significant improvement in tissue contrast, particularly the WM-GM contrast (4x in synthetic ULF data and 2.5x in 64mT data) against a minor trade-off ($\leq $4\%) in image quality metrics compared to other baselines. Our key contribution lies in contrast enhancement being solely driven by the physics of ULF image formation precluding the need for HF data compared to other ULF enhancement approaches \cite{LoHiResGAN,zhaolinUltraLowFieldMRIEnhancement2025}.

\par{
\textbf{Future directions: }
A natural extension of this work is to design an unsupervised segmentation tool for 64mT data. Although we generated target contrast vectors ($\mathbf{c}$) empirically from the LMIC dataset, we argue that $\mathbf{c}$ can be estimated using signal Bloch equations \cite{jogMRImageSynthesis2015,rousseauBrainHallucination2008}. 
In our future work, we will present a theoretical formulation for target contrast values and validate the modularity of this approach to other contrast types (T$_2$-weighted and PD-weighted images) along with designing a joint optimization strategy for ULF segmentation.
}

%
%
%

\bibliographystyle{splncs04}
\bibliography{references}

@inproceedings{blauPerceptionDistortionTradeoff2018,
  title = {The {{Perception-Distortion Tradeoff}}},
  booktitle = {2018 {{IEEE}}/{{CVF Conference}} on {{Computer Vision}} and {{Pattern Recognition}}},
  author = {Blau, Yochai and Michaeli, Tomer},
  year = {2018},
  month = jun,
  eprint = {1711.06077},
  primaryclass = {cs},
  pages = {6228--6237},
  doi = {10.1109/CVPR.2018.00652},
  urldate = {2026-04-16},
  archiveprefix = {arXiv}
}

@inproceedings{dayarathnaUltraLowFieldHighField2024,
  title = {Ultra {{Low-Field}} to {{High-Field MRI Translation Using Adversarial Diffusion}}},
  booktitle = {2024 {{IEEE International Symposium}} on {{Biomedical Imaging}} ({{ISBI}})},
  author = {Dayarathna, Sanuwani and Islam, Kh Tohidul and Chen, Zhaolin},
  year = {2024},
  month = may,
  pages = {1--4},
  issn = {1945-8452},
  doi = {10.1109/ISBI56570.2024.10635808},
  urldate = {2026-01-31}
}

@misc{diffusion_IQT,
  title = {A {{3D Conditional Diffusion Model}} for {{Image Quality Transfer}} -- {{An Application}} to {{Low-Field MRI}}},
  author = {Kim, Seunghoi and Tregidgo, Henry F. J. and Eldaly, Ahmed K. and Figini, Matteo and Alexander, Daniel C.},
  year = {2023},
  month = nov,
  number = {arXiv:2311.06631},
  eprint = {2311.06631},
  primaryclass = {cs, eess},
  publisher = {arXiv},
  doi = {10.48550/arXiv.2311.06631},
  urldate = {2024-07-25},
  archiveprefix = {arXiv}
}

@article{dohmenSimilarityQualityMetrics2025,
  title = {Similarity and Quality Metrics for {{MR}} Image-to-Image Translation},
  author = {Dohmen, Melanie and Klemens, Mark A. and Baltruschat, Ivo M. and Truong, Tuan and Lenga, Matthias},
  year = {2025},
  month = jan,
  journal = {Sci Rep},
  volume = {15},
  number = {1},
  pages = {3853},
  publisher = {Nature Publishing Group},
  issn = {2045-2322},
  doi = {10.1038/s41598-025-87358-0},
  urldate = {2025-10-23},
  copyright = {2025 The Author(s)},
  langid = {english}
}

@article{iglesiasQuantitativeBrainMorphometry2023,
  title = {Quantitative {{Brain Morphometry}} of {{Portable Low-Field-Strength MRI}}                     {{Using Super-Resolution Machine Learning}}},
  author = {Iglesias, Juan Eugenio et al. },
  year = {2023},
  month = mar,
  journal = {Radiology},
  volume = {306},
  number = {3},
  pages = {e220522},
  publisher = {Radiological Society of North America},
  issn = {0033-8419},
  doi = {10.1148/radiol.220522},
  urldate = {2025-10-27}
}

@misc{IXIDatasetBrain,
  title = {{IXI} {Dataset} - {Information eXtraction from Images. Biomedical Image Analysis Group, Imperial College London.}},
  urldate = {2025-10-16},
  langid = {american},
  url = {www.brain-development.org/ixi-dataset/}
}

@article{jogMRImageSynthesis2015,
  title = {{{MR}} Image Synthesis by Contrast Learning on Neighborhood Ensembles},
  author = {Jog, Amod and Carass, Aaron and Roy, Snehashis and Pham, Dzung L. and Prince, Jerry L.},
  year = {2015},
  month = aug,
  journal = {Medical Image Analysis},
  volume = {24},
  number = {1},
  pages = {63--76},
  issn = {1361-8415},
  doi = {10.1016/j.media.2015.05.002},
  urldate = {2026-01-30}
}

@inproceedings{lagunaSuperresolutionPortableLowfield2022,
  title = {Super-Resolution of Portable Low-Field {{MRI}} in Real Scenarios: Integration with Denoising and Domain Adaptation},
  shorttitle = {Super-Resolution of Portable Low-Field {{MRI}} in Real Scenarios},
  booktitle = {Medical {{Imaging}} with {{Deep Learning}}},
  author = {Laguna, Sonia et al.},
  year = {2022},
  month = apr,
  urldate = {2025-12-09},
  langid = {english}
}

@article{LoHiResGAN,
  title = {Improving Portable Low-Field {{MRI}} Image Quality through Image-to-Image Translation Using Paired Low- and High-Field Images},
  author = {Islam, Kh Tohidul and Zhong, Shenjun and Zakavi, Parisa and Chen, Zhifeng and Kavnoudias, Helen and Farquharson, Shawna and Durbridge, Gail and Barth, Markus and McMahon, Katie L. and Parizel, Paul M. and Dwyer, Andrew and Egan, Gary F. and Law, Meng and Chen, Zhaolin},
  year = {2023},
  journal = {Scientific Reports},
  volume = {13},
  number = {1},
  publisher = {Nature Publishing Group},
  issn = {2045-2322},
  doi = {10.1038/s41598-023-48438-1},
  langid = {english}
}

@article{lowfieldsim,
  title = {Minimum Field Strength Simulator for Proton Density Weighted {{MRI}}},
  author = {Wu, Ziyue and Chen, Weiyi and Nayak, Krishna S},
  year = {2016},
  month = may,
  journal = {PLOS ONE},
  volume = {11},
  number = {5},
  pages = {1--15},
  doi = {10.1371/journal.pone.0154}
}

@article{LowGAN,
  title = {Multi-Contrast High-Field Quality Image Synthesis for Portable Low-Field {{MRI}} Using Generative Adversarial Networks and Paired Data},
  author = {Lucas, Alfredo and Arnold, T. Campbell and Okar, Serhat V. and Vadali, Chetan and Kawatra, Karan D. and Ren, Zheng and Cao, Quy and Shinohara, Russell T. and Schindler, Matthew K. and Davis, Kathryn A. and Litt, Brian and Reich, Daniel S. and Stein, Joel M.},
  year = {2023},
  journal = {medRxiv},
  eprint = {https://www.medrxiv.org/content/early/2023/12/29/2023.12.28.23300409.full.pdf},
  publisher = {Cold Spring Harbor Laboratory Press},
  doi = {10.1101/2023.12.28.23300409},
  elocation-id = {2023.12.28.23300409}
}

@article{maLossOdysseyMedical2021,
  title = {Loss Odyssey in Medical Image Segmentation},
  author = {Ma, Jun and Chen, Jianan and Ng, Matthew and Huang, Rui and Li, Yu and Li, Chen and Yang, Xiaoping and Martel, Anne L.},
  year = {2021},
  month = jul,
  journal = {Medical Image Analysis},
  volume = {71},
  pages = {102035},
  issn = {1361-8415},
  doi = {10.1016/j.media.2021.102035},
  urldate = {2025-03-22}
}

@misc{molaeiImplicitNeuralRepresentation2023,
  title = {Implicit {{Neural Representation}} in {{Medical Imaging}}: {{A Comparative Survey}}},
  shorttitle = {Implicit {{Neural Representation}} in {{Medical Imaging}}},
  author = {Molaei, Amirali and Aminimehr, Amirhossein and Tavakoli, Armin and Kazerouni, Amirhossein and Azad, Bobby and Azad, Reza and Merhof, Dorit},
  year = {2023},
  month = jul,
  number = {arXiv:2307.16142},
  eprint = {2307.16142},
  primaryclass = {eess},
  publisher = {arXiv},
  doi = {10.48550/arXiv.2307.16142},
  urldate = {2025-03-30},
  archiveprefix = {arXiv}
}

@article{mr-physics-low-field,
  title = {Low-Field {{MRI}}: {{An MR}} Physics Perspective},
  author = {Marques, Jos{\'e} P. and Simonis, Frank F.J. and Webb, Andrew G.},
  year = {2019},
  journal = {Journal of Magnetic Resonance Imaging},
  volume = {49},
  number = {6},
  eprint = {https://onlinelibrary.wiley.com/doi/pdf/10.1002/jmri.26637},
  pages = {1528--1542},
  doi = {10.1002/jmri.26637}
}

@book{mri_book,
  title = {{{MRI}} from Picture to Proton},
  author = {McRobbie, Donald W. and Moore, Elizabeth A. and Graves, Martin J. and Prince, Martin R.},
  year = {2006},
  edition = {2},
  publisher = {Cambridge University Press},
  address = {Cambridge},
  doi = {10.1017/CBO9780511545405}
}

@misc{prob_deci_sim,
  title = {Deep Learning for Low-Field to High-Field {{MR}}: {{Image}} Quality Transfer with Probabilistic Decimation Simulator},
  author = {Lin, Hongxiang and Figini, Matteo and Tanno, Ryutaro and Blumberg, Stefano B. and Kaden, Enrico and Ogbole, Godwin and Brown, Biobele J. and D'Arco, Felice and Carmichael, David W. and Lagunju, Ikeoluwa and Cross, Helen J. and {Fernandez-Reyes}, Delmiro and Alexander, Daniel C.},
  year = {2019},
  eprint = {1909.06763},
  primaryclass = {eess.IV},
  archiveprefix = {arXiv}
}

@inproceedings{rousseauBrainHallucination2008,
  title = {Brain {{Hallucination}}},
  booktitle = {Proceedings of the 10th {{European Conference}} on {{Computer Vision}}: {{Part I}}},
  author = {Rousseau, Fran{\c c}ois},
  year = {2008},
  month = oct,
  series = {{{ECCV}} '08},
  pages = {497--508},
  publisher = {Springer-Verlag},
  address = {Berlin, Heidelberg},
  doi = {10.1007/978-3-540-88682-2_38},
  urldate = {2026-01-30},
  isbn = {978-3-540-88681-5}
}

@misc{saragadamWIREWaveletImplicit2023,
  title = {{{WIRE}}: {{Wavelet Implicit Neural Representations}}},
  shorttitle = {{{WIRE}}},
  author = {Saragadam, Vishwanath and LeJeune, Daniel and Tan, Jasper and Balakrishnan, Guha and Veeraraghavan, Ashok and Baraniuk, Richard G.},
  year = {2023},
  month = jan,
  number = {arXiv:2301.05187},
  eprint = {2301.05187},
  primaryclass = {cs},
  publisher = {arXiv},
  doi = {10.48550/arXiv.2301.05187},
  urldate = {2025-10-27},
  archiveprefix = {arXiv}
}

@inproceedings{shuaibuCapturingLongitudinalChanges2025,
  title = {Capturing {{Longitudinal Changes}} in {{Brain Morphology Using Temporally Parameterized Neural Displacement Fields}}.},
  booktitle = {Medical {{Imaging}} with {{Deep Learning}}},
  author = {Shuaibu, Aisha L. and Gibb, Kieran A. and Wijeratne, Peter A. and Simpson, Ivor J. A.},
  year = {2025},
  month = jan,
  urldate = {2025-03-30},
  langid = {english}
}

@misc{sitzmannImplicitNeuralRepresentations2020,
  title = {Implicit {{Neural Representations}} with {{Periodic Activation Functions}}},
  author = {Sitzmann, Vincent and Martel, Julien N. P. and Bergman, Alexander W. and Lindell, David B. and Wetzstein, Gordon},
  year = {2020},
  month = jun,
  number = {arXiv:2006.09661},
  eprint = {2006.09661},
  primaryclass = {cs},
  publisher = {arXiv},
  doi = {10.48550/arXiv.2006.09661},
  urldate = {2025-03-30},
  archiveprefix = {arXiv}
}

@article{smithFastRobustAutomated2002,
  title = {Fast Robust Automated Brain Extraction},
  author = {Smith, Stephen M.},
  year = {2002},
  month = sep,
  journal = {Hum Brain Mapp},
  volume = {17},
  number = {3},
  pages = {143--155},
  issn = {1065-9471},
  doi = {10.1002/hbm.10062},
  urldate = {2025-11-14},
  pmcid = {PMC6871816},
  pmid = {12391568}
}

@article{StochDecimSim,
  title = {Low-Field Magnetic Resonance Image Enhancement via Stochastic Image Quality Transfer},
  author = {Lin, Hongxiang and Figini, Matteo and D'Arco, Felice and Ogbole, Godwin and Tanno, Ryutaro and Blumberg, Stefano B. and Ronan, Lisa and Brown, Biobele J. and Carmichael, David W. and Lagunju, Ikeoluwa and Cross, Judith Helen and {Fernandez-Reyes}, Delmiro and Alexander, Daniel C.},
  year = {2023},
  month = jul,
  journal = {Medical Image Analysis},
  volume = {87},
  pages = {102807},
  publisher = {Elsevier BV},
  issn = {1361-8415},
  doi = {10.1016/j.media.2023.102807}
}

@article{super_res_Bouter,
  title = {Deep Learning-Based Single Image Super-Resolution for Low-Field {{MR}} Brain Images},
  author = {{de Leeuw den Bouter}, M. L. and Ippolito, G. and O'Reilly, T. P. A. and Remis, R. F. and {van Gijzen}, M. B. and Webb, A. G.},
  year = {2022},
  month = apr,
  journal = {Sci Rep},
  volume = {12},
  number = {1},
  pages = {6362},
  publisher = {Nature Publishing Group},
  issn = {2045-2322},
  doi = {10.1038/s41598-022-10298-6},
  urldate = {2024-07-30},
  copyright = {2022 The Author(s)},
  langid = {english}
}

@article{supervised_issues,
  title = {Challenges of Deep Learning in Medical Image Analysis -Improving Explainability and Trust},
  author = {Dhar, Tribikram and Dey, Nilanjan and Borra, Surekha and Sherratt, Robert},
  year = {2023},
  month = mar,
  journal = {IEEE Transactions on Technology and Society},
  volume = {PP},
  pages = {1--1},
  doi = {10.1109/TTS.2023.3234203}
}

@article{synthseg,
  title = {{{SynthSeg}}: {{Segmentation}} of Brain {{MRI}} Scans of Any Contrast and Resolution without Retraining},
  shorttitle = {{{SynthSeg}}},
  author = {Billot, Benjamin and Greve, Douglas N. and Puonti, Oula and Thielscher, Axel and Van Leemput, Koen and Fischl, Bruce and Dalca, Adrian V. and Iglesias, Juan Eugenio},
  year = {2023},
  month = may,
  journal = {Medical Image Analysis},
  volume = {86},
  pages = {102789},
  issn = {1361-8415},
  doi = {10.1016/j.media.2023.102789},
  urldate = {2024-07-30}
}

@misc{SynthSRPublicAI,
  title = {{{SynthSR}}: {{A}} Public {{AI}} Tool to Turn Heterogeneous Clinical Brain Scans into High-Resolution {{T1-weighted}} Images for {{3D}} Morphometry {\textbar} {{Science Advances}}},
  urldate = {2025-12-09},
  url = {www.science.org/doi/10.1126/sciadv.add3607}
}

@book{textbook_mri,
  title = {Magnetic {{Resonance Imaging}}: {{Physical Principles}} and {{Sequence Design}}},
  shorttitle = {Magnetic {{Resonance Imaging}}},
  author = {Brown, Robert W. and Cheng, Y.-C. Norman and Haacke, E. Mark and Thompson, Michael R. and Venkatesan, Ramesh},
  year = {2014},
  month = may,
  publisher = {John Wiley \& Sons},
  googlebooks = {rQGCAwAAQBAJ},
  isbn = {978-1-118-63397-7},
  langid = {english}
}

@misc{vandenbroekPaired64mT3T2025a,
  title = {Paired {{64mT}} and {{3T Brain MRI Scans}} of {{Healthy Subjects}} for {{Neuroimaging Research}}},
  author = {{van den Broek}, Ruben and Lena, Beatrice and {webb}, andrew},
  year = {2025},
  month = may,
  publisher = {Zenodo},
  doi = {10.5281/zenodo.15374450},
  urldate = {2025-10-16},
  langid = {english}
}

@article{wangDeepLearningImage2021,
  title = {Deep {{Learning}} for {{Image Super-Resolution}}: {{A Survey}}},
  shorttitle = {Deep {{Learning}} for {{Image Super-Resolution}}},
  author = {Wang, Zhihao and Chen, Jian and Hoi, Steven C. H.},
  year = {2021},
  month = oct,
  journal = {IEEE Transactions on Pattern Analysis and Machine Intelligence},
  volume = {43},
  number = {10},
  pages = {3365--3387},
  issn = {1939-3539},
  doi = {10.1109/TPAMI.2020.2982166},
  urldate = {2025-03-30}
}

@misc{wuIREMHighResolutionMagnetic2021a,
  title = {{{IREM}}: {{High-Resolution Magnetic Resonance}} ({{MR}}) {{Image Reconstruction}} via {{Implicit Neural Representation}}},
  shorttitle = {{{IREM}}},
  author = {Wu, Qing and Li, Yuwei and Xu, Lan and Feng, Ruiming and Wei, Hongjiang and Yang, Qing and Yu, Boliang and Liu, Xiaozhao and Yu, Jingyi and Zhang, Yuyao},
  year = {2021},
  month = jun,
  number = {arXiv:2106.15097},
  eprint = {2106.15097},
  primaryclass = {eess},
  publisher = {arXiv},
  doi = {10.48550/arXiv.2106.15097},
  urldate = {2025-03-25},
  archiveprefix = {arXiv}
}

@article{yeSuperResolutionBiomedicalImaging2023,
  title = {Super-{{Resolution Biomedical Imaging}} via {{Reference-free Statistical Implicit Neural Representation}}},
  author = {Ye, Siqi and Shen, Liyue and Islam, Md Tauhidul and Xing, Lei},
  year = {2023},
  month = oct,
  journal = {Phys Med Biol},
  volume = {68},
  number = {20},
  pages = {10.1088/1361-6560/acfdf1},
  issn = {0031-9155},
  doi = {10.1088/1361-6560/acfdf1},
  urldate = {2025-01-16},
  pmcid = {PMC10615136},
  pmid = {37757838}
}

@article{zhangSegmentationBrainMR2001,
  title = {Segmentation of Brain {{MR}} Images through a Hidden {{Markov}} Random Field Model and the Expectation-Maximization Algorithm},
  author = {Zhang, Y. and Brady, M. and Smith, S.},
  year = {2001},
  month = jan,
  journal = {IEEE Trans Med Imaging},
  volume = {20},
  number = {1},
  pages = {45--57},
  issn = {0278-0062},
  doi = {10.1109/42.906424},
  langid = {english},
  pmid = {11293691}
}

@InProceedings{zhaolinUltraLowFieldMRIEnhancement2025,
        author = { Islam, Kh Tohidul AND Ekanayake, Mevan AND Chen, Zhaolin},
        title = { { Ultra-Low-Field MRI Enhancement via INR-Based Style Transfer } },
        booktitle = {proceedings of Medical Image Computing and Computer Assisted Intervention -- MICCAI 2025},
        year = {2025},
        publisher = {Springer Nature Switzerland},
        volume = {LNCS 15975},
        month = {September},
        page = {597 -- 607}
}

\end{document}